\documentclass{INTERSPEECH2023}
\usepackage{todonotes}
\usepackage{comment}
\interspeechcameraready 

\title{External Language Model Integration for Factorized Neural Transducers}
\name{Michael~Levit, Sarangarajan~Parthasarathy, Cem~Aksoylar, Mohammad~Sadegh~Rasooli, Shuangyu~Chang}
\address{
  Microsoft, USA}
\email{mlevit|sarangp|ceaksoyl|mrasooli|shchang@microsoft.com}

\newcommand{\assgn}{=}
\def\rank{\mathop{\rm rank}}
\def\loglin{\mathop{\rm LLI}}
\def\lin{\mathop{\rm LI}}
\def\condlin1{\mathop{\rm C{\text -}LI}}
\def\sf{\mathop{\rm SF}}
\def\sm{\mathop{\rm \sigma}}
\def\lsm{\mathop{\rm log~\sigma}}

\def\blank{\mathop{\phi}}
\def\cname{$\langle$\textsc{name}$\rangle$ }
\def\ctype{$\langle$\textsc{type}$\rangle$ }
\def\NG{\mathop{\rm \scriptsize ng}}
\def\PT{\mathop{\rm \scriptsize pt}}

\begin{document}

\maketitle

\begin{abstract}
We propose an adaptation method for factorized neural transducers (FNT) with external language models. We demonstrate that both neural and n-gram external LMs add significantly more value when linearly interpolated with predictor output compared to shallow fusion, thus confirming that FNT forces the predictor to act like regular language models. Further, we propose a method to integrate class-based n-gram language models into FNT framework resulting in accuracy gains similar to a hybrid setup. We show average gains of 18\% WERR with lexical adaptation across various scenarios and additive gains of up to 60\% WERR in one entity-rich scenario through a combination of class-based n-gram and neural LMs.
\end{abstract}

\section{Introduction}
With large amounts of paired speech and text data available for ASR training, end-to-end (E2E) automatic speech recognition (ASR) systems have become competitive with hybrid ASR systems~\cite{sainath2020streaming,li2020developing,radford2022robust}. E2E alternatives have the advantage of architectural simplicity compared to hybrid ASR systems in which acoustic models, language models, and pronunciation models are independently trained. However, a consequence of joint modeling is that effective adaptation to target scenarios using text-only tuning data becomes challenging. There are generally four ways to perform text-only adaptation:
\begin{itemize}
\item {\bf Rescoring and reranking}: after decoding, use a powerful external LM to update scores and rerank n-best results or recognition lattice \cite{xu2022rescorebert,chiu2021innovative,beck2020lvcsr};
\item {\bf Audio generation}: generate audio for training texts via TTS \cite{deng2021improving, fazel2021synthasr}, data splicing \cite{zhao2021addressing} or learn to emulate encoder output from texts on the fly \cite{mittal2023in-situ};
\item {\bf Fusion and biasing}: during decoding interpolate posterior word probabilities with text priors from external LMs \cite{Chorowski2016Towards,gulcehre2017onusing,sriram2017coldfusion,toshniwal2018comparison, mcdermott2019density};
\item {\bf Explicit separation of internal LMs}: force the E2E decoder/predictor to behave more like a language model \cite{variani2020hybrid,meng2023modular,meng2021internal,zhou2022language, chen2022factorized}.
\end{itemize}
In this paper, we focus on the last approach using factorized neural transducers (FNT)~\cite{chen2022factorized} which is an extension of RNN-T~\cite{graves2012sequence} that makes the predictor network act like a proper language model. We investigate linear/log-linear interpolation approaches for augmenting FNT predictor activations with probabilities from external in-domain n-gram and neural models, and demonstrate that these methods are superior to the shallow fusion baseline. We also show how the n-gram approach can be extended to handle class-based n-gram LMs via linear interpolation, resulting in adaptation improvements comparable to those achievable with a hybrid recognizer. A number of important adjustments necessary for an effective application of the interpolated class-based setup is also discussed. 

The remainder of this paper is organized as follows. In \S\ref{sec:adapt} we explain the basic principles of FNTs and how various types of external LMs can be integrated with them. An experimental section follows, and the paper is concluded with a summary of presented results.

\section{LM Adaptation via Interpolation}\label{sec:adapt}
\begin{figure}[t]
  \centering
%   \scalebox{0.5}{
% \input{fnt_tikz}
% }
\includegraphics[width=0.45\textwidth]{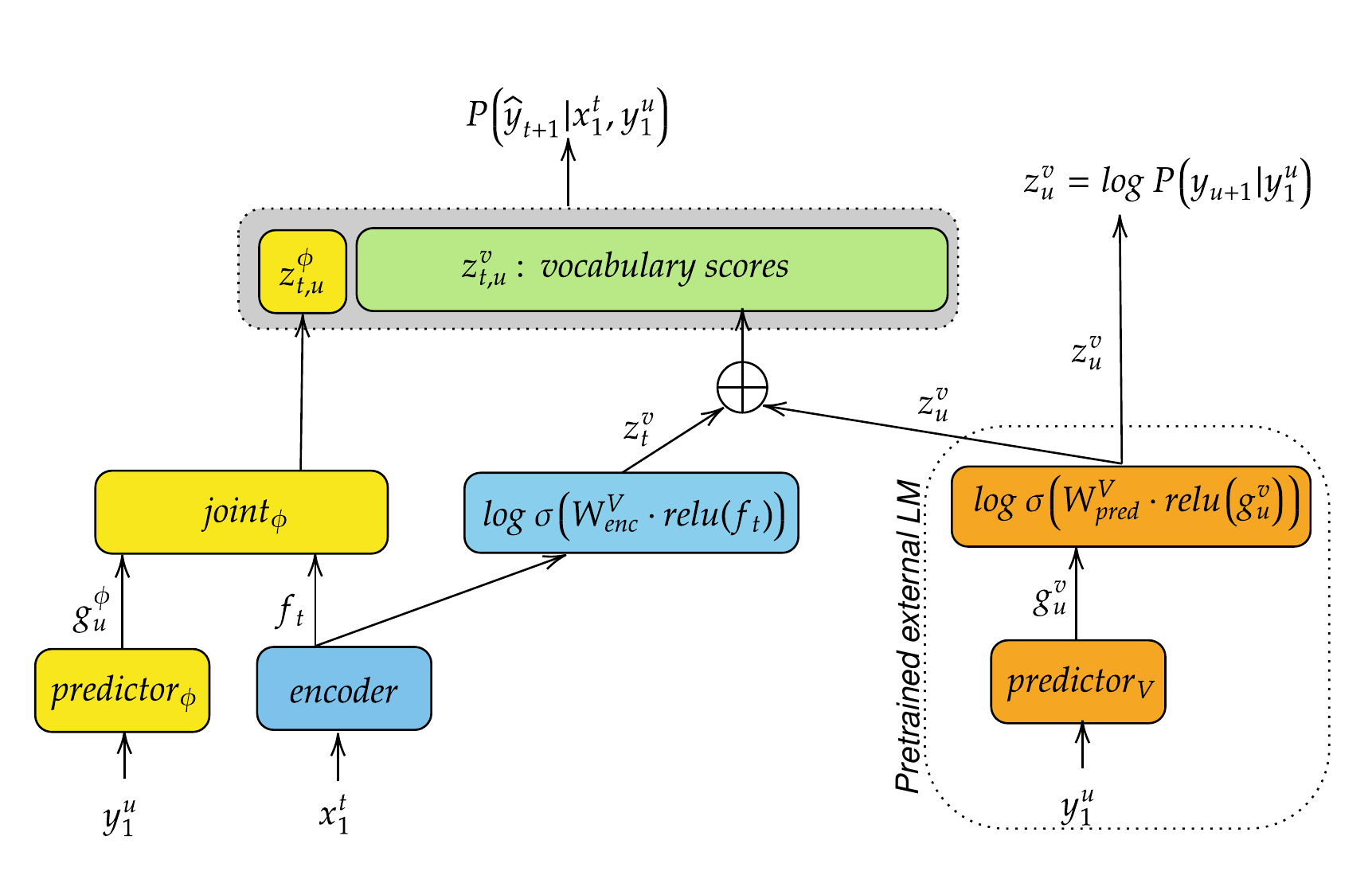}
  
  \caption{Schematic diagram of the factorized neural transducer model~\emph{\protect\cite{chen2022factorized}}. $\phi$ represents the \emph{blank} symbol.}
  \label{fig:fnt}
  \vspace{-1.8em}
\end{figure}

In this section, we describe our approach for LM adaptation via interpolation. The FNT model is an extension of the RNN-T model in which the predictor network is factored into two different predictors, one for the blank ($\blank$) likelihood and the other for non-blank words in vocabulary ${\cal V}$. We follow the model from \cite{chen2022factorized} which is depicted in Fig.~\ref{fig:fnt}. The following equations show the full pipeline of an FNT model (for the sake of brevity we omit biases):
\begin{eqnarray}
\pmb{f}_t &=& \mathrm{encoder}(x_1^t) \nonumber\\% ~~\in {\real}^{t\times1\times d_{f}}
\pmb{g}_u^{\blank} &=& \mathrm{predictor}_{\blank} (y_1^u)  \nonumber\\ %~~\in {\real}^{1\times u\times d_{g}}
\pmb{z}^{\blank}_{t,u} &=& W_{j} \cdot \mathrm{relu}( W_\mathrm{e}\cdot \pmb{f}_t + W_\mathrm{d}\cdot \pmb{g}_u^\phi)  \nonumber\\ %~~\in {\real}^{t\times u\times 1}
\pmb{z}^v_t&=&\lsm(\pmb{W}_\mathrm{enc}^v\cdot \pmb{f}_t)  \label{eq:fntenc}\\ % ~~\in {\real}^{t\times 1\times |{\cal V}|} 
\pmb{g}_u^v &=& \mathrm{predictor}_{\cal V} (y_1^u)  \nonumber\\ % ~~\in {\real}^{1\times u\times d_g}
\pmb{z}^v_u&=&\lsm(\pmb{W}_\mathrm{pred}^v\cdot \pmb{g}_u^v)  \label{eq:fntpred}\\ %~~\in {\real}^{1\times u\times |{\cal V}|}
\pmb{z}^v_{t,u}&=&\pmb{z}^v_t+\pmb{z}^v_u \label{eq:fntjoin}\\
P({\hat y}_{t+1}|x_1^t, y_1^u)&=&\sm([\pmb{z}^v_{t,u};\pmb{z}^{\blank}_{t,u}])\label{eq:fntsm} %~~\in {\real}^{t\times u\times |{\cal V}| + 1} 
\end{eqnarray}
where $\sigma$ is the softmax function. The values $P(\hat{y}_{t+1}|x_1^t, y_1^u)$ and $\pmb{z}^v_u$ are used for RNN-T and cross-entropy losses respectively during training and in some cases, $\pmb{z}^v_t$ can also be used for an additional CTC loss \cite{graves2006connectionist}.

Since the predictor in FNT behaves like an LM~\cite{chen2022factorized}, it can be adapted to the target domain by interpolating with an external LM~\cite{zhao2023fast}. This approach has several advantages over other approaches for E2E adaptation. For instance, an in-domain LM can be built on large text corpora and interpolated with the predictor in near-real time. Furthermore, unlike rescoring which takes place at the end of recognition,  interpolation can directly influence the beam search during decoding and preserve paths that could otherwise be dismissed in the beam. Finally, interpolation has leverage compared to shallow fusion in terms of explainability since the interpolation happens entirely within the space of prior probabilities. 

Language model fusion techniques \cite{Chorowski2016Towards,gulcehre2017onusing,sriram2017coldfusion} remain popular because they are simple to implement, reasonably effective, and generally robust.
We use shallow fusion (SF) \cite{Chorowski2016Towards} in our experiments as a baseline. Using notations above, Eqs.~(\ref{eq:fntpred},\ref{eq:fntjoin}) can be updated to account for SF, linear (LI) and log-linear (LLI) interpolation respectively:
\begin{eqnarray}
\bar{\pmb{z}}^v_{t,u}&=&\sf(\pmb{z}^v_{t,u}, \pmb{p}_u)\assgn\alpha \log\pmb{p}_u + (1-\alpha)\pmb{z}^v_{t,u} \label{eq:interpsf} \\
\bar{\pmb{z}}^v_u&=&\lin(\pmb{z}^v_u,\pmb{p}_u)\assgn\log(\alpha \pmb{p}_u + (1-\alpha)\exp(\pmb{z}^v_u)) \label{eq:interpli} \\
\bar{\pmb{z}}^v_u&=&\loglin(\pmb{z}^v_u,\pmb{p}_u)\assgn\alpha \log\pmb{p}_u + (1-\alpha)\pmb{z}^v_u \label{eq:interplli}
\end{eqnarray}
where $\pmb{p}_u$ is a tensor of word\footnote{For clarity, we will refer to basic modeling units as {\em words} and not {\em subwords} that they actually are, unless explicitly stated otherwise.} probabilities given history according to the external LM and $\alpha$ is interpolation weight assigned to the external model. If interpolation is applied to a subset of words rather than all of them, we can call it {\em biasing} \cite{aleksic2015bringing}. 

Combining predictor output with external neural LMs (NNLMs) is intuitive and simple. Assuming that the external model operates with the same lexicon (e.g. word-pieces) as the E2E network, $\pmb{p}_u$ is just a vector of normalized projections from the external NNLM. While used for RNN-T adaptation in several studies \cite{cabrera2021language,meng2023modular}, external NNLMs are difficult to estimate when only a small amount of text adaptation material is available. Instead, in such cases a large pretrained NNLM needs to be tuned on the in-domain data, which complicates personalization and plug-and-play abilities due to added time and memory requirements. With this caveat, we found that in some cases, and in particular for less-structured scenarios with a lot of conversational language, external NNLMs can offer superior adaptation results w.r.t. their n-gram counterparts. 

Integration of external NNLMs into FNT framework in both SF and interpolation settings is relatively straightforward and, for the most part, realized through augmentation of search states in sets $B$ (final states at each time step) and $A$ (expanded states) of the beam search decoder \cite{graves2012sequence} with hidden states and outputs of the external NNLM. 

\subsection{Adaptation with N-gram Language Models}\label{sec:anglm}
For n-gram models, LM states $s$ are short sequences ({\em histories}) of previous tokens. Therefore, definitions for RNN-T states in sets $A$ and $B$ have to be adjusted to contain these histories. However, querying n-gram language model probabilities is a one-by-one look-up process that breaks highly parallelizable tensor operations on neural networks. Additional optimization techniques such as one-step acceleration \cite{kim2020acceleraitng, kim2020acceleraitng2} also lose their attractiveness for n-gram adaptation for the same reason. 

To that end, we employ several optimization techniques to speed up interactions with n-grams LMs. In addition to extensive caching, we do not consult external n-gram LMs for all words in a given context but just for the ones with reasonably high probabilities (e.g. up to a certain \emph{rank} $r$ in a sorted list). Thus, for finite history $h$, external probability $\pmb{p}_u$ turns into a sparse vector of probabilities $P_{\NG}(w|h)$ that is defined only for some words $w$. This makes our approach similar to the biasing scheme used in \cite{zhao2019shallowfusion} except the decision which word to bias depends on its probability in the external LM and not on the word itself. The conditional linear interpolation formula (C-LI) for individual words of the $\bar{\pmb{z}}^v_u$ tensor can be written as: 
\begin{eqnarray}\label{eq:cond1}
    \bar{z}^v_u(w)&=&\condlin1(z^v_u(w), P_{\NG}(w|h))\\
    &\assgn& \left\{
    \begin{array}{ll}
       \lin(z^v_u(w), P_{\NG}(w|h)),&\mbox{if } \rank\limits_{P_{\NG}(w|h)}(w)<r\\
       z^v_u(w) & \mbox{otherwise}.\nonumber
    \end{array}
\right.
\end{eqnarray}

An efficient approximation for such conditional interpolation can be achieved by representing n-gram trie as a collection of sorted arrays \cite{heafield2011kenlm} such that for each history-node, identities and probabilities of all word-arcs can be found in constant time, low-probability words dismissed and iterative fall-backs to shorter histories performed until the requested number of high-probability words is collected. With this approximation and $r=200$, we observe only $\approx 20\%$ slow-down during inference due to interpolation with medium-sized n-grams models.

\subsection{Interpolating with Class-based LMs}
Class-based language models (CLMs) are much more difficult to work with in the context of neural E2E solutions. For simplicity, only the case of linear interpolation is considered here. In \cite{le2021deep}, SF biasing is done with a weighted FST representing one class embedded in characteristic carrier phrases. In contrast, we choose not to expand the class-based model into a single WFST, but preserve class definitions as prefix trees at the expense of more elaborate bookkeeping for CLM states $s$. With CLM trained on sentences tagged in terms of classes, its states $s$ now include not just previous token history $h$ (with tokens being words or classes), but also current position (node) $p$ in the prefix tree of the active class $c$. For example, in a phone assistant application with classes \cname (e.g. \emph{``john smith''}) and \ctype (e.g. \emph{``his mobile''}), a partial word hypothesis \emph{``$<$s$>$ call john smith on his…''} could end up in the following state $s$ of a 4-gram CLM:
\[
s=\{h:\text{\emph{$\langle$s$\rangle$ call \cname on}}, c:\text{\ctype}, p:\text{\emph{his}}\}
\]
The benefits of this representation include small memory footprint as well as absent need to  redistribute FSM weights \cite{pundak2018deep, le2021deep}, thanks to readily available cumulative posteriors in each node of the prefix tree. On the flip side, without deterministic WFST, we now face the ambiguity of having the same lexical history corresponding to potentially more than one state in the external CLM. For instance, the above lexical hypothesis could also be explained via state:
\[
s=\{h:\text{\emph{call \cname on his}}, c:\varnothing, p:\varnothing\}
\]
In \cite{Bruguier2022neural} a special {\em decider} network was trained to predict the next class given token history. Our CLM integration, on the other hand, relies entirely on beam search to find the most probable extension and does not require pretraining. 

\begin{figure}
    \centering
    \scalebox{0.75}{
    \input{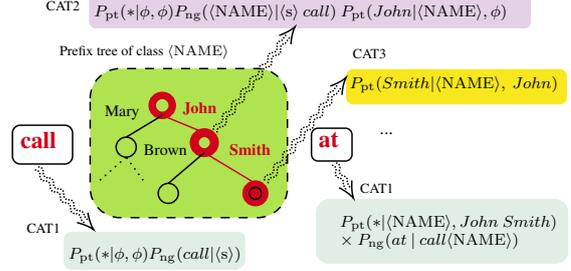}
    }
    \caption{Illustration of how three different transition types are applied to partial sentence ``call John Smith at\ldots''. }
    \label{fig:cat_fig}
\end{figure}

From the CLM point of view, RNN-T beam search expansions are mapped to three types of transitions between CLM states:
\begin{enumerate}
    \item[]\textbf{CAT1}: n-gram part of the CLM emits word $w$; % (shared lexicon with encoder/decoder)
    \item[]\textbf{CAT2}: prefix tree for class $c'$ is entered with word $w$;
    \item[]\textbf{CAT3}: word $w$ advances within the current class $c$.
\end{enumerate}
Let $P_{\PT}(*|c,p)$ be the exit probability from the current state $s=\{h,c,p\}$ inside the class $c$ (set to $1.0$ if $c=\varnothing$), and $s'=\{h', \varnothing, \varnothing\}$ the resulting new state. Then the CAT1/2/3 transition probabilities $P(w|s)$ are computed as:
\begin{eqnarray}
    P^1(w|s)&=&P_{\PT}(*|c,p)P_{\NG}(w|h')\\
    P^2(w|s)&=&P_{\PT}(*|c,p)P_{\NG}(c'|h')P_{\PT}(w|c',\varnothing)\\
    P^3(w|s)&=&P_{\PT}(w|c,p)
\end{eqnarray}
where $P_{\NG}(\cdot)$ values come from the CLM n-gram statistics, and $P_{\PT}(\cdot)$ from class prefix trees. Fig.~\ref{fig:cat_fig} shows an example of such transitions. A single word can trigger several CLM transition types but not all transitions types will be available for each word. To start, CAT2/3 transitions are only possible for words accepted in given class positions and also, to save computation, we will only consider CAT2/3 transitions for words %whose encoder probability is high. To express this in terms of rank $r$: $\rank\limits_{\pmb{z}^v_t(w)}(w) < r$. 
$w$ 
with high encoder scores: $\rank\limits_{\pmb{z}^v_t(w)}(w) < r'$.  
Given current LM state $s$, if $\mathcal{S}^1$, $\mathcal{S}^2$ and $\mathcal{S}^3$ are lists of words for which CAT1/2/3 transitions are possible (potentially with repetitions due to one-to-many mapping between word sequences and CLM states explained above), then we can construct an augmented version of $\pmb{z}^v_u$ as a concatenation of three parts:
\begin{eqnarray}\label{eq:czvu}
    \bar{\pmb{z}}^v_u&=&\Bigr[\left\{\condlin1(z^v_u(w), P^1(w|s))\ \ \forall w\in \mathcal{S}^1\right\},\\
                     & &~\ \left\{\lin(z^v_u(w),P^2(w|s))\ \ \forall w\in \mathcal{S}^2\right\},\nonumber\\
                     & &~\ \left\{\log(P^3(w|s))\ \ \forall w\in \mathcal{S}^3\right\}\Bigr]
                      \nonumber
\end{eqnarray}
Note that the within-class transition probabilities in CAT3 are accepted without interpolation with the predictor. The corresponding tensor of encoder probabilities is then constructed as:
\begin{eqnarray}\label{eq:czvt}
\bar{\pmb{z}}^v_t&=&\Bigr[\left\{z^v_t(w)\ \ \forall w\in \mathcal{S}^1\right\},\\
&&\ \ \ \left\{z^v_t(w)\ \ \forall w\in \mathcal{S}^2\right\}, \left\{z^v_t(w)\ \ \forall w\in \mathcal{S}^3\right\}\Bigr]\nonumber
\end{eqnarray}

The order of words in $\mathcal{S}^2$ and $\mathcal{S}^3$ needs to be observed in the same way in both Eqs.~(\ref{eq:czvu},\ref{eq:czvt}) and preserved throughout beam search iterations at each time step to select the best states for expansion. Typically, their sizes are much smaller than that of $\mathcal{S}^1$ and therefore the respective two components do not add a prohibitive amount of additional complexity to search. As a result, the average dimension of the augmented $\bar{\pmb{z}}^v_{t,u}=\bar{\pmb{z}}^v_t+\bar{\pmb{z}}^v_u$ increases only slightly, as does the search space for the next $y^*$ in terms of original notation in \cite{graves2012sequence}.

As for $\mathcal{S}^1$, it usually accounts for all words in ${\cal V}$ in the natural order, so that the first components in Eqs.~(\ref{eq:czvu},\ref{eq:czvt}) are simply $\pmb{z}^v_u$ and $\pmb{z}^v_t$. However, if $s$ is within a class $c$ and $P_{\PT}(*|c,p)=0$, then no CAT1 transitions are possible and $\mathcal{S}^1=\varnothing$. In this case, $\mathcal{S}^1$ is removed from the augmented definitions $\bar{\pmb{z}}^v_u$ and $\bar{\pmb{z}}^v_t$, leading to a speed-up in decoding. A further important refinement to this case concerns the situation where all three sets $\mathcal{S}^{1,2,3}$ are empty resulting in empty tensors in Eqs.~(\ref{eq:czvu},\ref{eq:czvt}). In absence of lexical competitors, emitting blank token $\phi$ would incur no costs.   
To alleviate this problem, we then compute the probability of $\phi$ by applying the final softmax Eq.~(\ref{eq:fntsm}) to a tensor of original values $\pmb{z}^v_{t,u}$ (before interpolation). Another potential way of addressing this situation is to properly separate blank probabilities in advance as suggested in \cite{variani2020hybrid, meng2023modular}.

Finally, to reduce the chance of having all search states in the beam traversing CLM entities only to discover a dead end, we modify the loop exit condition of the beam search algorithm to require at least some states in $B$ with available CAT1 options, effectively increasing average beam width by about 20\%.

\section{Experiments and Results}
\subsection{Scenarios and Recognition Setup}\label{sec:scen}
We evaluate our models on five English language setups covering a wide range of scenarios, with some setups containing multiple splits that we separately train and evaluate over: human-human instructions in spontaneous language (\textsc{HI}), human-human telephone conversations (\textsc{HTC} and \textsc{L-HTC}), strongly structured product ordering (\textsc{PO}) and human-machine interactive-voice-response systems (\textsc{HMIVR}). Each setup has a test set of about 1K utterances per split over which we report average WER, and training set sizes per split varying from $\sim$1.2K utterances for \textsc{PO} to $\sim$15K each for \textsc{HTC} and $\sim$24M for \textsc{L-HTC} as a notable exception. All data is anonymized.

Our baseline FNT system was trained on a large (though not production scale\footnote{We are not allowed to disclose exact size of the training material.}) anonymized corpus of paired audio and text data, compiled from a variety of Microsoft services. The encoder consists of 18 transformer layers and the predictor is a two-layer LSTM initialized by an LM trained on the text portion of the paired data. The word-piece \cite{kudo2018sentencepiece} vocabulary of 4.5K units was also trained on the same corpus as the pretrained LM. For more information on the training process please see \cite{Li2020high}. %, chen2022factorized}.

\subsection{Lexical Adaptation}

\begin{table*}[th]
  \caption{Baseline word error rate (WER) and word error rate reduction (WERR) due to different lexical adaptation methods}
  \label{tab:lexical}
  \centering
  \addtolength{\tabcolsep}{-2pt}
  \begin{tabular}{l|r|rrr|rrr|rrr|rrr}
    \toprule
    Corpus & Baseline & \multicolumn{3}{c}{NNLM WERR \scriptsize{($\alpha^*$)}} & \multicolumn{3}{c|}{N-gram WERR \scriptsize{($\alpha^*$)}} & \multicolumn{3}{c}{NNLM WERR \scriptsize{($\alpha^0$)}} & \multicolumn{3}{c}{N-gram WERR \scriptsize{($\alpha^0$)}} \\
           & WER      & SF & LI & LLI & SF & LI & LLI & SF \scriptsize{($0.05$)} & LI \scriptsize{($0.5$)} & LLI \scriptsize{($0.5$)} & SF \scriptsize{($0.25$)} & LI \scriptsize{($0.5$)} & LLI \scriptsize{($0.5$)} \\
    \midrule
    HI     & 31.5\% & 18\%  & \textbf{31\%}   & 28\%   & 11\%  & 25\%   & 21\%   & 13\%  & \textbf{29\%} & 25\% & 9\%   & 25\% & 19\% \\
    HTC    & 9.6\%  & 2\%   & \textbf{8\%}    & 6\%    & 1\%   & 5\%    & 3\%    &  2\%  &  \textbf{8\%} &  6\% & -4\%  &  3\% &  2\% \\
    L-HTC  & 15.6\% & 4\%   & \textbf{11\%}   & 9\%    & 1\%   & 7\%    & 5\%    &  4\%  & \textbf{11\%} &  9\% & 0\%   &  7\% &  5\% \\
    PO     & 21.2\% & 7\%   & 24\%   & 23\%   & 23\%  & \textbf{33\%}   & 29\%   &  4\%  & 23\% & 21\% & 18\%  & \textbf{31\%} & 25\% \\
    HMIVR  & 10.6\% & 1\%   & 8\%    & 14\%   & 5\%   & 29\%   & \textbf{41\%}   & -18\% & -8\% &  9\% & -11\% & 14\% & \textbf{41\%} \\
    \midrule
    Average & 16.3\% & 7.7\% & \textbf{17.8\%} & 16.0\% & 8.7\% & 17.6\% & 15.2\% & 4.5\%   & \textbf{16.0\%} & 14.3\% & 4.0\% & 15.7\% & 13.2\% \\
    \bottomrule
  \end{tabular}
\end{table*}

In this section, we evaluate adapted setups utilizing neural and n-gram LMs trained as described in \S\ref{sec:adapt} using lexical form adaptation training data for each scenario, comparing shallow fusion (\textsc{SF}, c.f.~Eq.~(\ref{eq:interpsf})) and interpolation (\textsc{LI/LLI}, Eqs.~(\ref{eq:interpli},\ref{eq:interplli})) approaches described in \S\ref{sec:adapt}.

Neural LMs adapted to the target scenario were trained as follows: we initialized the adapted LM by pretraining on a curated conversational corpus of about 10 billion words, froze the embedding parameters, and fine-tuned the parameters of the LSTM layers using CE loss for two passes over the adaptation data. For adaptation using n-gram LMs, we trained a 5-gram model with Kneser-Ney smoothing for each scenario, without utilizing any n-gram cutoffs or pruning. The adaptation data was represented using the same 4.5K word-piece units as the FNT model. The resulting model sizes ranged from $\sim$17K n-grams for \textsc{PO} scenarios to $\sim$600K for \textsc{HTC} scenarios, with the exception of $\sim$100M n-grams for \textsc{L-HTC} scenario.

For both n-gram and neural models, we evaluate with a fixed set of interpolation weights $\alpha \in \{0.01, 0.05, 0.1, $ $0.25, 0.5, 0.9\}$ where we omit the last two sets of weights for \textsc{SF} due to worsening performance. We present WERR performance for all three methods across all scenarios in Table~\ref{tab:lexical} for two weighting schemas: $\alpha^*$ where we choose the best-performing weight for each corpus-method combination and $\alpha^0$ where we choose a fixed weight for each method that performs best when averaged across all corpora. The best-performing weights for each method are denoted next to the method name. The bold items are the best-performing methods for each corpus and weighting scheme. The final averages are computed as a weighted average over the five setups.

From the results, we first observe that shallow fusion \textsc{SF} performs significantly worse than the other methods \textsc{LI/LLI}, whereas the two result in similar accuracy, with \textsc{LI} edging out \textsc{LLI} for all scenarios except \textsc{HMIVR}.
Another observation is that while NNLM-based adaptation gives slightly higher WERR than n-gram adaptation on average, there are certain scenarios where n-gram models significantly outperform, namely product ordering (\textsc{PO}) and human-machine IVR (\textsc{HMIVR}). We note that these two scenarios are very structured human-machine interactions, whereas the rest are more free-form and conversational. In particular, telephone conversations corpus (\textsc{HTC}) contains longer utterances that are less structured and consequently we see a larger gap between NNLM and n-gram adaptation, and lesser gains due to adaptation overall. Given the ease of training n-gram models on smaller amounts of data and the small observed performance gaps, n-gram adaptation is competitive with NNLM-based adaptation.
Finally, we note that the WERR performance over the space of $\alpha$'s is relatively smooth for non-\textsc{SF} methods, which can also be observed from the small WERR gap between two weight selection regimes $\alpha^*$ and $\alpha^0$. This is promising for practical applications since extensive tuning of weights does not seem to be necessary for good adaptation performance.

%For completeness, we also experimented with interpolating with the XLM without scenario-specific adaptation applied. This approach resulted in 3.4\% average WERR using the \textsc{LI} method with $\alpha^*$ weighting and 2.8\% WERR with $\alpha^0$. This demonstrates that significant gains are possible with XLM's even without scenario-specific adaptation data. 
%Finally, we also experimented with augmenting the n-gram models with the conversational data used to train the unadapted XLM before interpolating within the FNT framework which resulted in reduced WERR across all scenarios.\todo{Do we need to speculate on why?}

\subsection{Class-based Adaptation}
To test linear interpolation with external CLMs we chose one split of the product ordering scenarios \textsc{PO1} as well as two additional setups: one engineered data set of home entertainment command-and-control queries (\textsc{HE}) and another collection of personal assistant interactions mostly about calendar operations (\textsc{PA}). Tagged training sentences and class definitions for CLM have all been converted to word-pieces using the same model used to prepare FNT data. Table~\ref{tab:classb} offers more information regarding these corpora. All test sets contain $\sim$1200 utterances.

\begin{table}[th]
  \caption{Setups used for FNT+CLM evaluation}
  \label{tab:classb}
  \centering
  \begin{tabular}{llrr}
    \toprule
    Corpus & Train sent & Class sizes & Entity coverage \\
    \midrule
    PO1    &  35  & 1212, 121, 8, 8, 7, 4 & ~50-80\% \\
    HE     & 731  & 827 & 100\%\\
    PA     & 15 & 990 & ~80\%\\
    \bottomrule
  \end{tabular}
  \vspace{-1em}
\end{table}

The effect of incorporating CLM via linear interpolation with FNT predictor is summarized in Table~\ref{tab:clm}. A very significant WER improvement of up to 55\% is observed on all domains. At this point, the optimal interpolation weights $\alpha^*$ have been found via parameter sweep. However, here too, the weight space appears to be smooth and one should be able to optimize a single weight parameter with a small validation set. In addition to FNT numbers, we also present results for respective hybrid ASR setups where word-level CLM was interpolated with the large n-gram LM with $\alpha$ optimized similarly. It can be seen that FNT and hybrid WER improvements are in the same ballpark, providing yet another confirmation of a proper LM separation in FNT. Note that comparison of absolute WER numbers does not make sense in context of these setups, because FNT and hybrid ASR were trained on different training data.

\begin{table}[th]
  \caption{Effect of linear interpolation with CLM}
  \label{tab:clm}
  \centering
  \begin{tabular}{l|lll|ll}
    \toprule
    Corpus & \multicolumn{3}{c|}{FNT} & \multicolumn{2}{c}{Hybrid}\\
           & Baseline & $\alpha^*_{\mbox{\scriptsize FNT}}$  & WERR & $\alpha^*_{\mbox{\scriptsize hyb}}$ & WERR \\
    \midrule
    PO1    & 21.1\%   & 0.9         & 54\% &  0.2       & 31\%       \\
    HE     & 5.3 \%   & 0.6         & 55\% &  0.9       & 73\%       \\
    PA     & 20.6\%   & 0.9         & 36\% &  0.9       & 60\%       \\
    \bottomrule
  \end{tabular}
\end{table}

Since for \textsc{PO1} we also had a separate lexical training corpus as described in \S\ref{sec:scen}, a three-way interpolation between predictor, external adapted NNLM and CLM was possible. We applied it in consecutive manner, first linearly interpolating with NNLM, and then with the CLM using previously found optimal interpolation weights. The outcome was further reduced WER on this set with overall WERR reaching {\bf 60\%}.

\section{Conclusions}
We showed how external language models can be integrated into FNT framework. Our experiments proved that in-domain n-gram and neural LMs can be linearly interpolated with the predictor model and improve WER far beyond what is possible with shallow fusion. In addition, we suggested an extension of the integration algorithm that also allows using class-based n-gram models for linear interpolation, achieving WERR of over 50\% on our internal sets. A combination of neural and class-based n-gram LMs produced the best overall results.

\bibliographystyle{IEEEtran}
\bibliography{mybib}

\end{document}